\documentclass{llncs} %

\usepackage{float}

\usepackage{booktabs}
\usepackage{graphicx}
\usepackage{wrapfig}
\usepackage{comment}

\usepackage[labelfont=bf]{caption}
\usepackage[colorlinks=true,linkcolor=blue,citecolor=blue]{hyperref}

\usepackage{color,soul}
\usepackage{amsmath} 
\usepackage{mathtools} 
\usepackage{amsfonts} 
\usepackage{amssymb}

\usepackage{array}
\usepackage{booktabs}
\usepackage{rotating}
\usepackage{multirow}
\usepackage{multicol}
\usepackage{lscape}
\usepackage{subfig}
\usepackage{times}

\usepackage[export]{adjustbox}

\usepackage[ruled,vlined]{algorithm2e}
\usepackage{xargs}                      

\usepackage[colorinlistoftodos,prependcaption,textsize=tiny]{todonotes}

\usepackage{color}
\definecolor{reddish}{rgb}{0.82, 0.1, 0.26}

\usepackage{tikz,xcolor,hyperref}

\definecolor{lime}{HTML}{A6CE39}
\DeclareRobustCommand{\orcidicon}{
	\begin{tikzpicture}
	\draw[lime, fill=lime] (0,0) 
	circle [radius=0.16] 
	node[white] {{\fontfamily{qag}\selectfont \tiny ID}};
	\draw[white, fill=white] (-0.0625,0.095) 
	circle [radius=0.007];
	\end{tikzpicture}
	\hspace{-2mm}
}

\foreach \x in {A, ..., Z}{\expandafter\xdef\csname orcid\x\endcsname{\noexpand\href{https://orcid.org/\csname orcidauthor\x\endcsname}
			{\noexpand\orcidicon}}
}
			
\begin{document}

\title{Deep Graph Normalizer: A Geometric Deep Learning Approach for Estimating Connectional Brain Templates }

\titlerunning{Short Title}  

\author{Mustafa Burak Gurbuz\index{Gurbuz, Mustafa Burak}  \and Islem Rekik\orcidA{} \index{Rekik, Islem}\thanks{ {corresponding author: irekik@itu.edu.tr, \url{http://basira-lab.com}. }}}

\institute{BASIRA Lab, Faculty of Computer and Informatics, Istanbul Technical University, Istanbul, Turkey}

\authorrunning{M.B Gurbuz et al.}  

\maketitle              

\begin{abstract}
A connectional brain template (CBT) is a normalized graph-based representation of a population of brain networks —also regarded as an `average' connectome. CBTs are powerful tools for creating representative maps of brain connectivity in typical and atypical populations. Particularly, estimating a well-centered and representative CBT for populations of \emph{multi-view} brain networks (MVBN) is more challenging since these networks sit on complex manifolds and there is no easy way to fuse different heterogeneous network views. This problem remains unexplored with the exception of a few recent works rooted in the assumption that the relationship between connectomes are mostly linear. However, such an assumption fails to capture complex patterns and non-linear variation across individuals. Besides, existing methods are simply composed of sequential MVBN processing blocks without any feedback mechanism, leading to error accumulation. To address these issues, we propose Deep Graph Normalizer (DGN), \emph{the first geometric deep learning (GDL) architecture} for normalizing a population of MVBNs by integrating them into a single connectional brain template. Our end-to-end DGN learns how to fuse multi-view brain networks while capturing non-linear patterns across subjects and preserving brain graph topological properties by capitalizing on graph convolutional neural networks. We also introduce a randomized weighted loss function which also acts as a regularizer to minimize the distance between the population of MVBNs and the estimated CBT, thereby enforcing its \emph{centeredness}. We demonstrate that DGN significantly outperforms existing state-of-the-art methods on estimating CBTs on both small-scale and large-scale connectomic datasets in terms of both representativeness and discriminability (i.e., identifying distinctive connectivities fingerprinting each brain network population). Our DGN code is available at \url{https://github.com/basiralab/DGN}\footnote{\textbf{Paper YouTube video:} \url{https://www.youtube.com/watch?v=Q_WLY2ZNxRk&t=3s}}\footnote{\textbf{Code demo video:} \url{https://www.youtube.com/watch?v=8tDNYV0NYpY&t=1s}}.
	
\end{abstract}

\keywords{Connectional brain templates $\cdot$ Deep graph normalizer $\cdot$ Population multiview brain network integration}

\section{Introduction}


The field of network neuroscience has made substantial advances in characterizing the human brain network (or connectome modeling the pairwise relationship between brain regions of interest (ROIs)) by means of large-scale connectomic datasets collected through various projects such as Human Connectome Project (HCP) \cite{HCP} and Connectome Related to Human Disease (CRHD) \cite{CRHD}. These rich and multimodal brain datasets allow us to map brain connectivity and efficiently detect atypical deviations from the healthy brain connectome. Particularly, learning how to \emph{normalize} a population of brain networks by estimating a \emph{well-centered} and \emph{representative} connectional brain template (CBT) is an essential step for group comparison studies as well as discovering the integral signature of neurological disorders \cite{Dhifallah:2019}. Intuitively, a CBT can be defined as a `normalized connectome' of a population of brain networks, which can be simply produced by linear averaging. However, such a normalization technique is very sensitive to outliers and overlooks non-linear relationships between subjects. This normalization process can be regarded as an `integration' or `standardization' of brain networks. 


More broadly, estimating a CBT of a population of heterogeneous multi-view brain networks, where each view captures particular traits of the brain construct (e.g., cortical thickness, function, cognition), is even a more challenging task since such connectomic data might lie on complex high-dimensional manifolds. To address this challenge, \cite{Dhifallah:2019} proposed a clustering-based approach based on similarity network fusion (SNF) \cite{SNF} to fuse multi-view brain networks in each cluster. Fused networks are then linearly averaged to produce a CBT for a population of multi-view brain networks (MVBN). Despite its promising results, \cite{Dhifallah:2019} heavily depends on the selection of the number of clusters. In order to overcome this problem, \cite{Dhifallah:2020} introduced the netNorm framework, which instead of clustering, constructs a high-order graph using cross-view connectional features as nodes and their Euclidean distance as dissimilarity measure to select the most centered brain connections in a population of MVBN. Next, the selected connections are integrated into a single network using SNF \cite{SNF}. Currently, netNorm presents the state-of-the-art method by outperforming SCA \cite{Dhifallah:2019} and other baseline methods in the CBT estimation task. However, netNorm \cite{Dhifallah:2020} has several limitations. \emph{First}, it uses  Euclidean distance as a \emph{pre-defined metric} for selecting the most representative brain connection which might fail to capture complex non-linear patterns in the brain connectome across subjects. \emph{Second}, netNorm also uses SNF for fusing different views. Even though SNF is a powerful tool since it is a generic unsupervised technique, it comes with assumptions such as emphasizing top $k$ local connections for each node (i.e., brain region) and equally averaging the global topology of complementary networks for each iterative update to ultimately merge them. Instead of relying on such general assumptions, this MVBN normalization process can instead be \emph{learned} to decide which information provided by the networks is important for the target CBT estimation. \emph{Third}, netNorm consists of \emph{independent} feature extraction, feature selection, and fusion steps. These fully independent steps cannot provide feedback to each other in order to globally optimize the CBT estimation process. Therefore errors might accumulate throughout the estimation pipeline.

To address all these limitations, we propose \textbf{Deep Graph Normalizer (DGN)}: an unprecedented approach capitalizing on geometric deep learning (GDL) to learn how to normalize a population of heterogeneous MVBNs and generate a well-representative and centered CBT in an \emph{end-to-end} fashion. Although GDL achieved remarkable results in several recent biomedical data analysis works such as disease classification \cite{classification} and protein interface prediction \cite{protein}, to the best of our knowledge, no previous works used GDL to address the problem of integrating a population of multi-view networks \cite{survey1,survey2}. To fill this gap, we present several major contributions to the state-of-the-art  as follows. \emph{First}, we design a GDL architecture that takes multi-view brain networks and maps them into a normalized CBT. Specifically, we propose a GDL architecture that maps MVBN of a training subject to a population-representative CBT.  Brain networks of each training subject passes through several layers of graph convolutional neural networks that are consecutively applied to learn hidden embeddings for each node (i.e., brain ROI) by locally integrating connectivities offered by different heterogeneous views and blending the previous layer's embeddings using integrated connectivities. Next, we compute the pairwise absolute difference of the final layer's node embeddings to derive connectivity weights of the generated CBT. \emph{Second}, we introduce the Subject Normalization Loss (SNL) which is a randomized weighted loss function that evaluates the representativeness of a generated \emph{subject-biased} CBT (i.e. obtained by feeding a particular subject to the model) against a random subset of brain networks in the training set to achieve \emph{subject-to-population mapping}. Besides forcing the model to learn how to generate population-based representative CBTs by fusing complementary information supplied by MVBNs, SNL also acts as a regularization due to randomization and different weights assigned to each view according to their connectivity weight distributions. \emph{Third}, finalized CBT can be obtained by feeding an arbitrary subject of the training population to the trained model since the model learns how to map any subject to a population-representative CBT thanks to SNL optimization. However, the choice of the subject biases the output CBT and leads to non-optimal performance. We introduce a post-training step to overcome this bias and further refine the finalized CBT. 

\begin{figure}[ht!]
\centering
{\includegraphics[width=12.5cm]{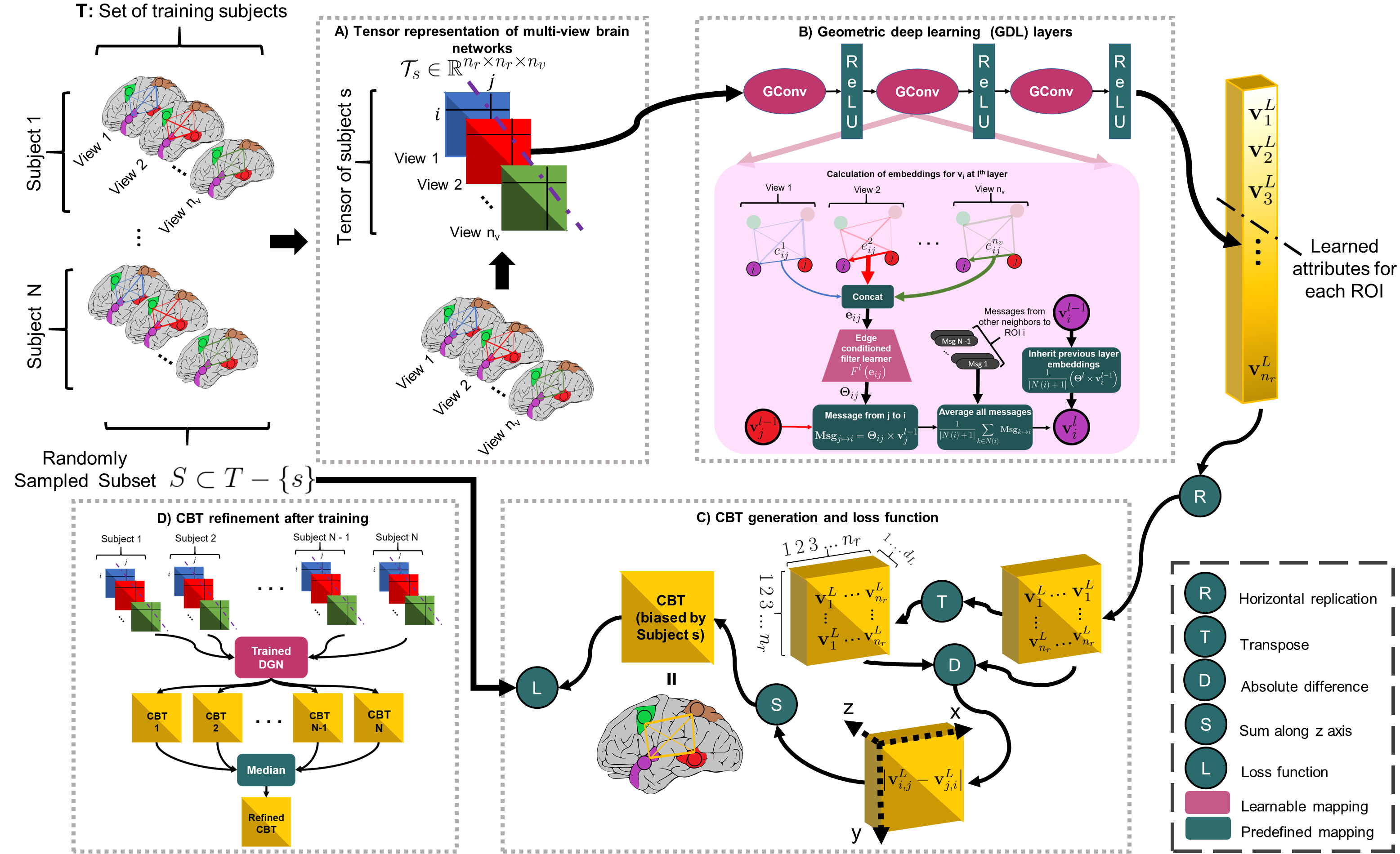}}
\caption{\emph{Proposed Deep Graph Normalizer (DGN) architecture for estimating connectional brain templates for a given population of multi-view brain networks.} \textbf{(A) Tensor representation of multi-view brain networks.} Each subject $s$ is represented by $\mathcal{T}_{s} \in  \mathbb{R}^{n_{r} \times n_{r} \times n_{v}}$, composed of a set of undirected, fully connected graphs, each capturing single connectional feature. \textbf{(B) Geometric deep learning layers.} Our model includes a sequence of edge conditioned \cite{edge-conv} graph convolutional neural network layers which are separated by ReLU non-linearity. Each layer learns deeper embeddings for ROIs by utilizing activation of the previous layer and topological structure of the brain network. \textbf{(C) CBT generation and loss function.} ROI embeddings that are output by the final layer are passed through a series of tensor operations to calculate the pairwise absolute difference of each pair of nodes for CBT construction. Next, the representativeness of the estimated CBT is evaluated against a random subset of training views for loss calculation. \textbf{(D) CBT refinement after training.} To select the most centered connections for final CBT generation, we first pass each training subject through the trained model to generate its corresponding CBT. Finally, we produce the final CBT by selecting the element-wise median of  all training CBTs. }

\label{fig:1}
\end{figure}

\section{Proposed Method}

In this section, we detail the components of our DGN architecture for estimating CBTs (\textbf{Fig.}~\ref{fig:1}). First, each subject in a population of multi-view brain networks is represented by an undirected fully-connected graph where each node (i.e., brain ROI) is initialized with identity features (trivially set to 1) and each edge has $n_{v}$ attributes that correspond to connectivity weights in different network views. We generate CBTs using a training set of MVBNs $T = \{ \mathbf{T}_1^1, \mathbf{T}_2^1, \dots, \mathbf{T}_i^v, \dots, \mathbf{T}_N^{n_v}  \}$ , where $\mathbf{T}_{i}^{v}$ denotes the $v^{th}$ brain view of subject $i$, and evaluate the representativeness on a testing set using 5-fold cross-validation. These training subjects pass through 3 layers of GDL layers that includes edge conditioned filter learner (a shallow neural network). From layer to layer, deeper embeddings are learned for each ROI using edge-conditioned graph convolution \cite{edge-conv} which aggregates the information passed by its neighbours while taking into consideration the multi-view attributes of its neighboring edges. We then use ROI embeddings output by the final layer to produce the connectivity matrix of the generated CBT by calculating the between the final embeddings of each ROI pair.  This generated CBT is evaluated against a random subset of the training MVBNs for loss optimization. Once the DGN architecture is fully trained using this randomized training sample selection strategy, each training subject is then fed to the trained model to produce a population representative CBTs that are \emph{biased by the given input subject}. Finally, we eliminate outlier connectivities due to subject bias by taking the element-wise median of all possible CBTs. We detail all these steps in what follows.


\textbf{A- Tensor representation of multi-view brain networks}. Given a population of subjects, we represent each subject $s$ by a tensor $\mathcal{T}_{s} \in  \mathbb{R}^{n_{r} \times n_{r} \times n_{v}}$ (\textbf{Fig.}~\ref{fig:1}--A) that is composed by stacking connectivity matrices of $n_{v}$ brain networks where each network has $n_{r}$ nodes (i.e., ROI). Note that using this subject-specific brain tensor representation, an  $\mathbf{e}_{ij} \in \mathbb{R}^{n_v \times 1}$ connecting ROIs $i$ and $j$ encapsulates $n_{v}$ attributes. We set diagonal entries in the tensor to zero to eliminate self-loops (i.e., ROI self-connectivity). In addition to inputting edge multi-view attributes, our DGN architecture also takes a node attributes matrix $\mathbf{V}^{0} \in \mathbb{R}^{n_{r} \times d_{0}}$ as an input, where $d_{0}$ denotes the number of initial attributes for each ROI. As for each brain ROI, we do not have predefined attributes, we set each entry of the $\mathbf{V}^{0}$ matrix to `$1$' (i.e., identity), and through our deep model training for optimizing the SNL function, we  learn these node-specific implicit attribute representations by simply using the input edge attributes. In each graph convolution layer and for each multi-view brain connection $\mathbf{e}_{ij}$ linking nodes $i$ and $j$,  we generate an edge-specific weight matrix utilizing an \emph{edge-conditioned filter learner} \cite{edge-conv}. Then these weight matrices multiplied by ROIs attributes to compute the next layer's ROI attributes therefore after the first convolution each ROI will have a different set of attributes even though they were identical in the beginning.

\textbf{B- Geometric deep learning layers}. In this section, we detail the graph convolutional layers of our architecture that maps ROIs with identical attributes to a high dimensional distinctive representations by utilizing edge features $\mathbf{e}_{ij}$ between ROIs. This is achieved through 3 graph convolutional network layers (\textbf{Fig.}~\ref{fig:1}--B) with  edge-conditioned convolution operation \cite{edge-conv} that are separated by ReLU non-linearity. Each layer $l\in \{ 1, 2, ..., L \}$  includes edge-conditioned filter learner neural network $F^{l} : \mathbb{R}^{n_v} \mapsto  \mathbb{R}^{d_{l} \times d_{l - 1}}$ that dynamically generates edge specific weights for filtering message passing between ROIs $i$ and $j$ given the features of $\mathbf{e}_{ij}$. This operation is defined as follows:

\begin{gather*}
 \mathbf{v}{_{i}^{l}} =    \mathbf{\Theta}^{l}  .\mathbf{v}{_{i}^{l-1}} +  \frac{1}{\left |N(i)\right |} \left (\sum_{j \epsilon N(i)} \ F^{l}(\mathbf{e}_{ij}; \mathbf{W}^{l}) \mathbf{v}^{l - 1}_{j} + \mathbf{b}^{l}\right );  F^{l}(\mathbf{e}_{ij}; \mathbf{W}^{l}) = \mathbf{\Theta}_{ij}
\end{gather*}

where $\mathbf{v}{_{i}^{l}}$ is the embedding of ROI $i$ at layer $l$, $\mathbf{\Theta}^{l}$ is a learnable parameter, and $N(i)$ denotes the neighbours of ROI $i$. $\mathbf{b}^{l} \in \mathbb{R}^{d_{l}}$ denotes a network bias and $F^{l}$ is a neural network that  maps $\mathbb{R}^{n_{v}}$ to $\mathbb{R}^{d_{l} \times d_{l - 1}}$ with weights $\mathbf{W}^{l}$. $\mathbf{\Theta}_{ij}$ represents the dynamically generated edge specific weights by $F^{l}$. Note that $F^{l}$ can be any type of neural network and vary in each layer depending on the characteristics and complexity of edge weights.

\textbf{C-  CBT construction layer and subject normalization loss function}. 
After obtaining the output  $\mathbf{V}^{L} = \left [ \mathbf{v}_1^L, \mathbf{v}_2^L, ..., \mathbf{v}_{n_{r} - 1}^L, \mathbf{v}_{n_{r}}^L \right ]^T$ of the final DGN layer, which consists of embeddings for each ROI, we compose the output CBT by computing the pair-wise absolute difference of the learned embeddings. We formulate this process using several tensor operations (\textbf{Fig.}~\ref{fig:1}--C) for easy and efficient backpropagation.  First, $\mathbf{V}^{L}$  is replicated horizontally $n_{r}$ times to obtain $\mathcal{R} \in \mathbb{R}^{n_{r} \times  n_{r} \times  d_{L}}$. Next, $\mathcal{R}$ is transposed (replacing all elements $\mathcal{R}_{xyz}$ with $\mathcal{R}_{yxz}$)    to get $\mathcal{R}^{T}$. Last, we compute the element-wise absolute difference of $\mathcal{R}$ and $\mathcal{R}^{T}$. The resulting tensor is summed along $z$-axis (i.e. size of node embeddings) to estimate the final CBT $\mathbf{C} \in \mathbb{R}^{n_{r} \times  n_{r}}$. 

\emph{Subject Normalization Loss (SNL) optimization}. We propose to evaluate the representativeness of the generated CBT using a random subset of the training subject views. This random selection procedure has two main advantages compared to evaluating against all training subjects. First, randomization has a regularization effect since it is much easier for the model to overfit if the loss is calculated against the same set of subjects in each iteration. Secondary, the sample size can be fixed to a constant number so that the magnitude of the loss and the computation time will be independent of the size of the training set. Note that since SNL compares generated CBT with a subset of training subject views, model weights updated in a way so that the generated CBT represents a population of MVBNs even though it is rooted in a single subject input. Given the generated CBT $\mathbf{C}_{s}$ for subject $s$ and a random subset $S$ of training subject indices, we define our SNL for training subject $s$ and the optimization loss as follows:
\begin{equation*}
SNL_{s} = \sum_{v = 1}^{n_{v}} \sum_{i \in S} \left \|  \mathbf{C}_{s} - \mathbf{T}_{i}^{v} \right \|_{F} \times \lambda_{v} \textcolor{blue}{;} \
\min\limits_{\mathbf{W}_1, \mathbf{b}_1 \dots \mathbf{W}_L, \mathbf{b}_L}  \frac{1}{\left | T \right |} \sum_{s = 1}^{\left | T \right |}\\ SNL_{s}
\label{eq:1}
\end{equation*}
The $\lambda_{v}$ is a view specific normalization term that is defined as:
\begin{equation*}
	\lambda _{v} = \frac{\frac{1}{\mu_{v}}}{\max \left \{   \frac{1}{\mu_{j}} \right \}_{j = 1}^{n_{v}}}
\end{equation*}
where $\mu_{v}$ is the mean of brain graph connectivity weights of view $v$ and $\max \left \{  \frac{1}{\mu_j} \right \}_{j = 1}^{n_{v}}$ is the maximum of mean reciprocals $\frac{1}{\mu_{1}}$ to $\frac{1}{\mu_{n_v}}$. We use this view-specific normalization weight since brain network connectivity distribution and value range might largely vary across views. This will help avoid view-biased CBT estimation where the trained model might overfit some views and overlook others. Related problems in the literature are addressed by normalizing the connectivity matrix. For example, SNF \cite{SNF} divides connectivities in each row by the sum of the entries in that row to normalize measurements; however, this breaks the symmetry in the views, therefore, it is not applicable in our case.  Another simple normalization approach such as using min-max scaling can saturate our inputs at 0 and 1 while standard z-score scaling generates negative connectivities in the graph that is not suitable for representing a fully positive brain connectomes (such as structural or morphological). Therefore, we introduce $\lambda _{v}$ to ensure that the model gives equal attention to each brain view regardless of their value range.

\textbf{D- CBT refinement after the training}. Our model learns to map multi-view brain networks of a particular subject $s$ to a population-based representative CBT $\mathbf{C}_{s}$. Although all of the CBTs generated by the model are representative of the randomly sampled training population, they are biased towards the given input subject $s$. To eliminate this bias, we propose an additional step (\textbf{Fig.}~\ref{fig:1}--D) to obtain a more refined and representative CBT for the whole training set. First, each subject is fed through the trained model to obtain corresponding CBTs. Next, most centered connections are selected from these CBTs by calculating the element-wise median. The median operation produces a valid view with a non-negative symmetrical adjacency matrix. This operation could also be replaced with other measures of central tendency; however, we used the representativeness score to verify that the median is the most suitable for our case.

\section{Results and Discussion}

\textbf{Connectomic datasets and model hyperparameter setting.} We benchmarked our DGN against state-of-the-art method netNorm for CBT estimation \cite{Dhifallah:2020} on  two small-scale and large-scale connectomic datasets using 5-fold cross-validation. The first dataset (AD/LMCI dataset) consists of 77 subjects (41 subjects diagnosed with Alzheimer's diseases (AD) and 36 with Late
Mild Cognitive Impairment (LMCI)) from the Alzheimer's Disease Neuroimaging Initiative (ADNI) database GO public dataset \cite{ADdataset}. Each subject is represented by 4 cortical morphological brain networks derived from maximum principal curvature, the mean cortical thickness, the mean sulcal depth, and the average curvature as in \cite{Raeper:2018,Lisowska:2019,Nebli:2019}. The second dataset (NC/ASD dataset) is collected from the Autism Brain Imaging Data Exchange ABIDE I public dataset \cite{ASDdataset} and includes 310 subjects (155 normal control (NC) and 155 subjects with autism spectral disorder (ASD)) with 6 cortical morphological brain networks extracted from the 4 aforementioned cortical measures in addition to cortical surface area and minimum principle area. For each hemisphere, the cortical surface is reconstructed from T1-weighted MRI using FreeSurfer pipeline \cite{FreeSurfer} and parcellated into 35 ROIs using Desikan-Killiany atlas \cite{parc} and its corresponding brain network is derived by computing the pairwise absolute difference in cortical measurements between pairs of ROIs. 

We trained 8 different models to generate CBTs for both hemispheres of 4 populations namely; AD, LMCI, NC, and ASD. We empirically set all hyperparameters for the DGN models using grid search. Each model includes 3 edge-conditioned convolutional neural network layers with an edge-conditioned filter learner neural network that maps 4  (for AD/LMCI dataset) or 6 (for NC/ASD dataset) attributes obtained from heterogeneous views to $\mathbb{R}^{d_{l} \times d_{l - 1}}$. These layers are separated by ReLU activation function and output embeddings with 36, 24 and 5 (for AD/LMCI dataset) or 8 (for NC/ASD dataset) dimensions for each ROI in the MVBN, respectively. We trained all models using gradient descent with Adam optimizer with a learning rate of $0.0005$. We fixed the number of random samples in our SNL function to $10$.

\begin{figure}[htp!]
\centering
\includegraphics[width=12cm]{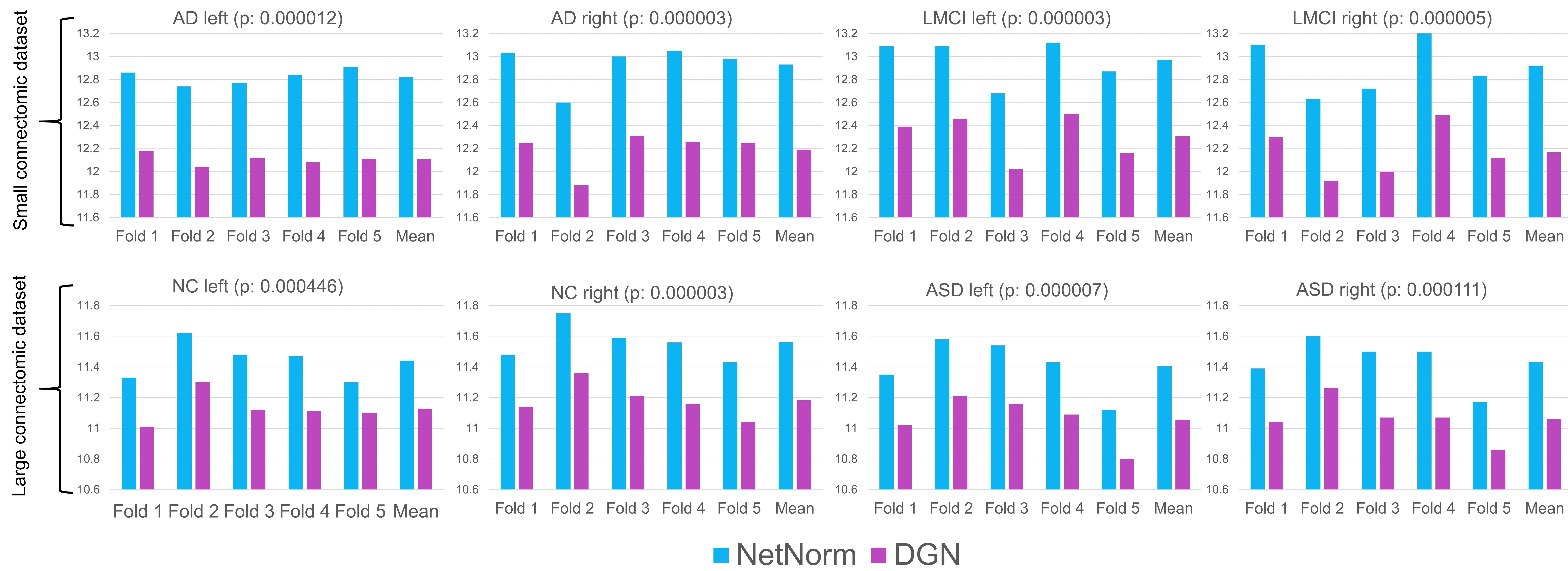}
\caption{\emph{Representativeness comparison between CBTs generated by the proposed model and netNorm \cite{Dhifallah:2020}.}
	Charts illustrate the average Frobenius distance between the CBTs generated using the training set and the network views in the testing set. Also, p-values obtained by two-tailed t-test are reported for each population. LH: left hemisphere. RH: right hemisphere.}
\label{fig:3}
\end{figure}

\textbf{CBT representativeness test.} To evaluate the representativeness of generated CBT, we computed mean Frobenius distance which is calculated as $d_{F}(A,B) = \sqrt{\sum_{i} \sum_{j}  \left | A_{ij} - B_{ij} \right |^{2} }$ between the estimated CBT and the different views in the testing set.  We split both datasets into training and testing sets using 5-fold cross-validation for reproducibility and generalizability. \textbf{Fig.}~\ref{fig:3} depicts the average Frobenius distance between CBTs generated by DGN and netNorm \cite{Dhifallah:2020} using the training set and the views in the left out test population.  We note that our proposed model significantly ($p < 0.001$) outperforms netNorm in terms of representativeness across all left-out folds and evaluation datasets.

\textbf{CBT discriminativeness reproducibility test.} We hypothesize that a well-representative CBT can capture the most discriminative traits of a population of MVBNs, acting as a connectional brain fingerprint. To test this hypothesis, we first spot the top $k$ most discriminative brain connectivities where a class-A CBT largely differs from a class-B CBT. To do so, we compute the absolute difference between both estimated CBTs using respectively A and B populations. Next, we sum the columns of this difference matrix to obtain discriminability score for each brain ROI. We then pick the top $k$ most discriminative ROIs with the highest score. To evaluate the reproducibility of CBT-based discriminative ROIs, for each brain view $v$, we independently train a support vector machine (SVM) and with a supervised feature selection  method. Specifically, we extract connectional features  from each brain network view by vectorizing the upper triangle entries. Next, for each network view, we use 5-fold randomized partitioning to divide each population $p^{A}$ and $p^{B}$  into 5 subpopulations. For each brain view $v$ and a combination of $p^{A}_{i}$ and $p^{B}_{j}$ a SVM is trained and a weight vector $\mathbf{w}_{ij}^{v}$ that scores the discriminativeness of each feature (i.e., ROI)  is learned using Multiple Kernel Learning (MKL), which is a wrapper method assigning weights to features according to their distinctiveness for the given classification task. The final feature weight vector is computed by summing up the weight vectors for all views and all possible A-B combinations of their 5 subpopulations as follows: $\mathbf{\omega} = \sum_{\mathbf{w} = 1}^{n_{v}} \sum_{i,j = 1}^{5} \mathbf{w}_{i,j}^{v}$. Next, we anti-vectorize  $\mathbf{\omega}$ vector and obtain matrix $ \mathbf{M} \in \mathbb{R}^{n_{r} \times n_{r}}$. By summing the columns of the resulting matrix we get ROIs discriminability scores. Finally, we picked the top $k$ ROIs with the highest score. \textbf{Table}~\ref{tab:1} reports the overlap between the top $k=15$ ROIs identified using CBT-based method (netNorm and DGN) and MKL-based SVM method. Remarkably, our proposed model not only generates more representative and centered CBTs but is significantly more reproducible in discriminability than netNorm \cite{Dhifallah:2020}. 

\textbf{Discovery of most discriminative brain ROIs for each disordered population.} DGN also revealed left insula cortex, left superior temporal sulcus (STS) and right frontal pole as most discriminative regions of ASD population, which resonates with existing findings on autism. {\cite{insula} reports that the alteration of the left insula in the ASD population might be the cause of abnormalities in emotional and affective functions. Next, by comparing activation of STS in different social scenarios, \cite{STS} shows that the dysfunction of STS is the essential factor of social perception impairment in autism. For instance, in contrast to NC subjects, individuals with autism show hypoactivation in the STS when exposed to matched visual and auditory information. Furthermore, \cite{STS} demonstrates that healthy children had greater response in STS triggered by biological (e.g. human movement) than by non-biological motions (e.g. clock). However, STS activation's of children with autism do not differ significantly depending on the nature of the motion. Lastly, \cite{pole} shows that the faces of boys with ASD have atypical right dominant asymmetry and suggests that the asymmetric growth of the right frontal pole can explain this facial anomaly. As for the AD-LMCI dataset, DGN picked the left temporal pole (TP) and right entorhinal cortex (EC) as the most discriminative regions of the brain. \cite{TL} highlights that the pathological changes in TP are a common trait among all AD patients. Moreover, \cite{EC} confirms that the alteration of the EC is a good biomarker of AD and LMCI and indicates that the AD patients show greater atrophy in the right EC which supports DGN's choice.

\begin{table}
	\centering
	\begin{tabular}{c c c c}
		\hline\noalign{\smallskip}
		Overlap Rate & netNorm \cite{Dhifallah:2020}  & \bf{DGN}  \\
		\hline\noalign{\smallskip}
		AD-LMCI Left Hem. & 0.60 & \bf{0.73} \\
		AD-LMCI Right Hem. & 0.33 & \bf{0.40} \\
		\hline\noalign{\smallskip}
	\end{tabular}
	\begin{tabular}{c c c c}
		\hline\noalign{\smallskip}
		Overlap  Rate & netNorm \cite{Dhifallah:2020}  & \bf{DGN}  \\
		\hline\noalign{\smallskip}
		NC-ASD Left Hem. & 0.53 & \bf{0.53} \\
		NC-ASD Right Hem. & 0.33 & \bf{0.40} \\
		\hline\noalign{\smallskip}
	\end{tabular}
	\caption{\emph{Overlap rate between ROIs selected by MKL and CBT-based methods.}}
\label{tab:1}
\end{table}

\section{Conclusion}

In this paper, we introduced Deep Graph Normalizer for estimating connectional brain templates for a given population of multi-view brain networks. Beside capturing non-linear patterns across subjects, the proposed method also learns how to fuse complementary information offered by MVBNs in an end-to-end fashion. We showed that the proposed DGN outperformed state-of-the-art method for estimating CBTs in terms of both representativeness and discriminability. In our future work, we will evaluate our architecture on multi-modal brain networks such as functional and structural brain networks while capitalizing on  geometric deep learning for estimating \emph{holistic} CBTs. Also, we will introduce topological loss constraints such as  Kullback-Leibler divergence of node degree distributions of generated CBTs and population brain networks  to further ensure that the generated CBTs are topologically sound.


\section{Acknowledgments}

I. Rekik is supported by the European Union's Horizon 2020 research and innovation programme under the Marie Sklodowska-Curie Individual Fellowship grant agreement No 101003403 (\url{http://basira-lab.com/normnets/}).

\bibliography{Biblio3}
\bibliographystyle{splncs}
\end{document}